\documentclass[letterpaper]{article} 
\usepackage{aaai2026}  
\usepackage{times}  
\usepackage{helvet}  
\usepackage{courier}  
\usepackage[hyphens]{url}  
\usepackage{graphicx} 
\usepackage{natbib}  
\usepackage[colorlinks=true, allcolors=blue]{hyperref}

\usepackage{caption} 
\usepackage{algorithm}
\usepackage{algorithmic}

\usepackage{newfloat}
\usepackage{listings}
\DeclareCaptionStyle{ruled}{labelfont=normalfont,labelsep=colon,strut=off} 
\floatstyle{ruled}
\newfloat{listing}{tb}{lst}{}
\floatname{listing}{Listing}
\usepackage{algorithm}
\usepackage{algorithmic}
\usepackage{xspace}
\usepackage{enumitem}
\usepackage{amsmath}
\usepackage{comment}
\usepackage{booktabs}
\usepackage{multirow}
\usepackage{amssymb}
\usepackage[most]{tcolorbox}
\usepackage{tabularx}
\usepackage{newfloat}
\usepackage{listings}
\newcommand{\tool}{Agent-SAMA\xspace}
\newcommand{\phead}[1]{\noindent {\bf #1}}

\title{\tool: State-Aware Mobile Assistant}
\author{
    Linqiang Guo\textsuperscript{\rm 1},
    Wei Liu\textsuperscript{\rm 1}\thanks{Corresponding author.},
    Yi Wen Heng\textsuperscript{\rm 1},
    Tse-Hsun (Peter) Chen\textsuperscript{\rm 1}\thanks{Principal investigator.},
    Yang Wang\textsuperscript{\rm 2}
}
\affiliations{
    \textsuperscript{\rm 1}Software PErformance, Analysis, and Reliability (SPEAR) lab, Concordia University, Montreal, Canada\\
    \textsuperscript{\rm 2}Concordia University, Canada\\

    g\_linqia@live.concordia.ca, w\_liu201@encs.concordia.ca, he\_yiwen@encs.concordia.ca, peterc@encs.concordia.ca, yang.wang@concordia.ca
}

\makeatletter
\renewcommand{\copyright@text}{}
\def\@copyrightspace{}
\makeatother

\begin{document}

\maketitle

\begin{abstract}
Mobile Graphical User Interface (GUI) agents aim to autonomously complete tasks within or across apps based on user instructions. While recent Multimodal Large Language Models (MLLMs) enable these agents to interpret UI screens and perform actions, existing agents remain fundamentally reactive. They reason over the current UI screen but lack a structured representation of the app navigation flow, limiting GUI agents’ ability to understand execution context, detect
unexpected execution results, and recover from errors. We introduce \tool, a state-aware multi-agent framework that models app execution as a Finite State Machine (FSM), treating UI screens as states and user actions as transitions. \tool implements four specialized agents that collaboratively construct and use FSMs in real time to guide task planning, execution verification, and recovery.
We evaluate \tool on two types of benchmarks: cross-app (Mobile-Eval-E, SPA-Bench) and mostly single-app (AndroidWorld). On Mobile-Eval-E, \tool achieves an 84.0\% success rate and a 71.9\% recovery rate. On SPA-Bench, it reaches an 80.0\% success rate with a 66.7\% recovery rate. Compared to prior methods, \tool improves task success by up to 12\% and recovery success by 13.8\%. On AndroidWorld, \tool achieves a 63.7\% success rate, outperforming the baselines.
Our results demonstrate that structured state modeling enhances robustness and can serve as a lightweight, model-agnostic memory layer for future GUI agents.
\end{abstract}

\begin{links}
     \link{Code}{https://doi.org/10.5281/zenodo.15430187}
 \end{links}

\section{Introduction}
\label{sec:introduction}
Mobile apps have become an important part of people's daily life, providing access to online shopping, communication, social media, and more. 
To automate and support these interactions, recent research has introduced graphical user interfaces (GUI) agents that are powered by Multimodal Large Language Models (MLLMs)~\cite{MobileGPT, AutoDroid, wang2024mobile_agent, yang2023appagent, zhang2024_mobileexperts, li2024_appagentv2, wang2024mobile_agent_v2, wang2025mobile}. 
Given a user instruction, such as ``\textit{using Chrome to search for the date of the next Winter Olympics opening ceremony, and then setting a reminder for that date in
Calendar}'', GUI agents can automatically complete these tasks without human intervention. The agents can analyze and reason app UI screens, identify actionable UI elements (e.g., buttons, input fields), and perform actions such as tapping, typing, or scrolling~\cite{yang2023appagent, wang2024mobile_agent, wang2024mobile_agent_v2, wang2025mobile}. By chaining these interactions together, GUI agents can complete complex multi-step tasks across diverse app environments.

\begin{figure*}
  \centering
  \includegraphics[width=0.9\linewidth]{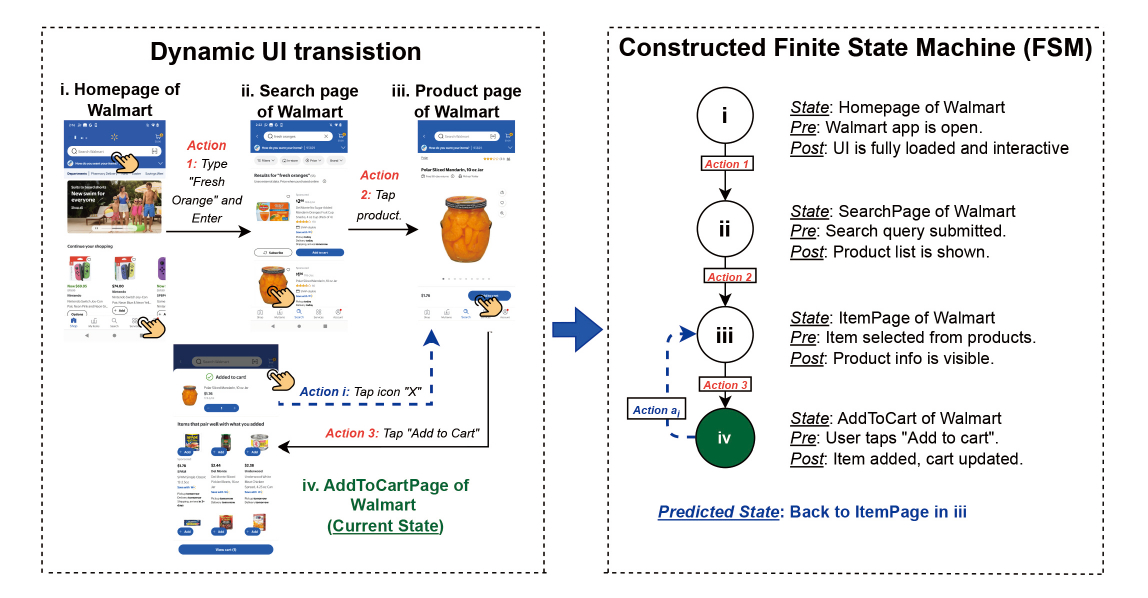}
  \caption{An example of how \tool represents real-time UI interactions as a Finite State Machine (FSM). The left side shows the dynamic UI transitions of the Walmart App along with the user action (e.g., typing and tapping) that leads to a new UI screen. The right side shows the corresponding FSM, where each UI screen is represented as a state (a natural language description of the screen generated by MLLM) with its MLLM-generated pre- and post-condition. The user action defines the transition between the states. Given the current state and the entire FSM, \tool also predicts the possible next state. }
  \label{fig:fsm}
\end{figure*}

However, existing GUI agents remain fundamentally reactive: they primarily reason the next action based on the current UI screen, without maintaining a structured representation of app behaviors—\textit{like tourists who navigate a city one street at a time, knowing where they have been but lack a coherent view of the route or how different locations are connected}. 
This lack of structural understanding of an app’s navigation logic limits GUI agents’ ability to interpret execution context, detect unexpected execution outcomes, and recover from errors. 
To illustrate, Figure~\ref{fig:fsm} shows how a user interacts with the Walmart app by performing a sequence of actions (e.g., tapping), where each action may lead to a transition to a new state (i.e., a new UI screen).
The user types a query to search for products, selects a product from the results to view its details on \textit{product page}, and taps the button to add the product to the shopping cart. Then, the user can tap ``X'' to return to the previously visited \textit{product page}. 

In app usages, every step depends and builds on the prior one to follow a task-oriented logical flow according to the app's requirements and design. 
Understanding such interaction flows is essential. As illustrated in the right side of Figure~\ref{fig:fsm}, app usage is not a series of isolated actions. Instead, the transitions between UI screens are triggered by specific user actions, and follow a pre-defined and structured changes in UI states. Having structured representations of the transition provides several benefits for GUI agents: it enables the agent to track its progress within a task, anticipate the outcomes of actions, verify whether the resulting state aligns with expectations, and assist in finding a recovery state among the action sequences. 

In this paper, we propose \textbf{\tool, a novel mobile GUI agent framework that leverages finite state machines (FSMs) to represent structured and state-aware task executions}. 
An FSM is a formal computational model that represents systems as a finite number of states with state transitions triggered by specific actions. 
Given that mobile apps are inherently stateful systems, where each UI screen corresponds to a distinct state and user actions trigger transitions, FSMs provide a natural abstraction for app usages~\cite{IEEE_Hierarchical_sm_for_app,wang2004using, wagner2006fsm, Espada_2015_SM, shahbaz2022fsmapp}.

Specifically, \tool consists of four phases: 1) \textbf{Planning}: given a task, the \textbf{Planner Agent} decomposes the task into a sequence of subtasks. To improve the planning process, \tool generates several candidate plans and applies LLM-as-judges to select the best among all~\cite{gu2025surveyllmasajudge}. 
2) \textbf{Execution}: After planning, \tool guides subtask executions through the collaboration of three agents. The \textbf{Screen Parser} extracts structured information from the current UI screen, generating a description of the screen and UI element locations (e.g., buttons). The output is passed to the \textbf{State Agent}, which incrementally constructs an FSM in real-time to model the navigation flow by mapping each screen's natural language description to a distinct state and associating user actions with transitions. State Agent adds pre- and post-conditions to every state and transition, allowing for better reasoning and verification. Finally, the \textbf{Actor Agent} selects and performs the appropriate action (e.g., tapping at a specific button) based on the output from the State Agent. 3) \textbf{Error Recovery and Verification}: If an unexpected error happens, the \textbf{Reflection Agent} compares the FSM-predicted transition (including post-conditions) against the current screen and uses the FSM to identify a previously verified stable state. It then generates a recovery plan to return to that point for retry. 4) \textbf{Knowledge Retention}: After completing a task, the \textbf{Mentor Agent} stores the constructed FSMs and executed actions in \tool's long-term memory storage to create useful knowledge that can guide future task executions.

Experimental results show that \tool achieves significant improvement over the baseline Mobile-Agent-E+Evo~\cite{wang2025mobile} across two challenging cross-app benchmarks~\cite{wang2025mobile, chen2025spabench} and also achieves a 63.7\% success rate on AndroidWorld~\cite{rawles2025androidworld}. 
On Mobile-Eval-E~\cite{wang2025mobile}, \tool improved the Success Rate by over 12\% compared to the baseline, completing more tasks successfully. It also encounters less error during task execution (32 vs. 49) and yet achieves a 4.53\% higher Recovery Success rate. The results highlight \tool's ability to detect and recover from execution errors during runtime. We observe similar trends on SPA-Bench~\cite{chen2025spabench}, where \tool outperforms the baseline by 5\% in Success Rate and a 13.81\% improvement in Recovery Success rate with less errors (63 vs. 70). On AndroidWorld, \tool outperforms several baselines such as Mobile-Eval-E+Evo~\cite{wang2025mobile} and AgentS2~\cite{agashe2025agents2}, indicating its strong performance through stable planning and execution.

\noindent\textbf{Our contributions are as follows:}
\begin{itemize}[leftmargin=12pt]
    \item \textbf{Introducing FSM Modeling to Mobile GUI Agents.} We are the first to incorporate finite state machine (FSM) modeling into mobile GUI agents. By treating each app as a state machine—with UI screens as states and user actions as transitions—we enable structured, state-aware reasoning that addresses key limitations of reactive execution. Source code and evaluation data are available at Zenodo ~\cite{zenodo}.
    \item \textbf{A State-Aware Agent Framework with Persistent Memory and Recovery.} We implement FSM modeling in \tool, a mobile GUI agent that constructs per-app state graphs during execution, tracks visited states, infers state preconditions and postconditions, and uses this structured representation for proactive planning and robust error recovery.
    \item \textbf{An LLM-Based Judge for Plan Selection and Replanning.} To enhance planning and decision-making, we introduce a LLM–based judge that evaluates multiple candidate plans and selects the most reliable one based on context and execution history. This mechanism further improves the agent's ability to adapt and recover in complex mobile environments. 
    \item \textbf{Improved Task Performance Across Benchmarks.} \tool achieves significant performance gains over the baseline, Mobile-Agent-E + Evo, on both Mobile-Eval-E and SPA-Bench benchmarks, including up to 12\% improvement in Success Rate, over 6.5\% in Action Accuracy, over 7\% in Satisfaction Score, and up to 13.81\% in Error Recovery Rate. 
\end{itemize}

\section{Related Work}
\label{sec:related_work}
\subsection{Mobile GUI Agents}
Recent advances in MLLM have significantly improved GUI agents' ability in executing complex tasks. Some agents adopt a single-agent architecture (e.g., {Mobile-agent}~\cite{wang2024mobile_agent}, {AppAgent}~\cite{yang2023appagent}, {AppAgent v2}~\cite{li2024_appagentv2}), while others follow a multi-agent paradigm (e.g., {MobileExperts}~\cite{zhang2024_mobileexperts}, {Mobile-Agent-v2}~\cite{wang2024mobile_agent_v2} and {Mobile-Agent-E}~\cite{wang2025mobile}) that distributes perception, planning, and execution across specialized agents. 
To improve task success rates, most existing mobile agents~\cite{MobileGPT, AutoDroid, wang2024mobile_agent, yang2023appagent, zhang2024_mobileexperts, li2024_appagentv2, wang2024mobile_agent_v2, wang2025mobile} rely on prompt engineering~\cite{liu2025llmpoweredguiagentsphone}, enriching LLM input with UI information (e.g., screenshots, UI trees, OCR), short action histories, and chain-of-thought (CoT) reasoning. 
A few studies also explore training-based approaches, including supervised fine-tuning~\cite{ding2024_mobileagent} and reinforcement learning~\cite{liu2024_autoglm}. 

One similar study, {GUI-Xplore}~\cite{sun2025gui}, builds GUI transition graphs from offline videos to support static reasoning tasks such as screen recall. While both GUI-Xplore and \tool use graph-based representations of app navigation, GUI-Xplore focuses on evaluating agents' ability to reason about app structure without real-time interaction or execution. In contrast, \tool constructs FSMs online during task execution.  
Using the constructed FSM, \tool integrates state tracking and reasoning, transition validation, and recovery planning into the agent's decision-making process. Our FSM-based architecture is model-agnostic and compatible with existing agent designs, offering a potential avenue for improving future mobile GUI agents.

\subsection{Evaluation Benchmarks for GUI Agents}
Numerous benchmarks have been proposed to evaluate mobile GUI agents, including MobileEnv~\cite{DanyangZhang2023_MobileEnv}, AutoDroid~\cite{2024_AutoDroid_benchmark}, MobileAgentBench~\cite{wang2024_mobileagentbench}, AndroidArena~\cite{2024_AndroidArena_benchmark}, and  AndroidWorld~\cite{rawles2025androidworld}. However, many of these fall short in realism, long-horizon task modeling, or cross-app execution. For example, MobileEnv and AndroidWorld rely on emulator-based environments. MobileAgentBench mainly targets single-app interactions, while AndroidArena
supports some cross-app tasks but still lacks long-horizon or highly complex workflows. In this work, We adopt Mobile-Eval-E~\cite{wang2025mobile} and SPA-BENCH~\cite{chen2025spabench}, which provide 364 actions and 262 actions (in English version apps), respectively. These benchmarks support realistic, multi-step, and cross-application tasks, enabling rigorous evaluation of our proposed {Agent-SAMA}. They also facilitate direct comparison with the baseline, Mobile-Agent-E~\cite{wang2025mobile}, particularly for complex behavior modeling.

\section{\tool}
\label{sec:agent_sama}
\begin{figure*}
  \centering
\includegraphics[width=1\linewidth]{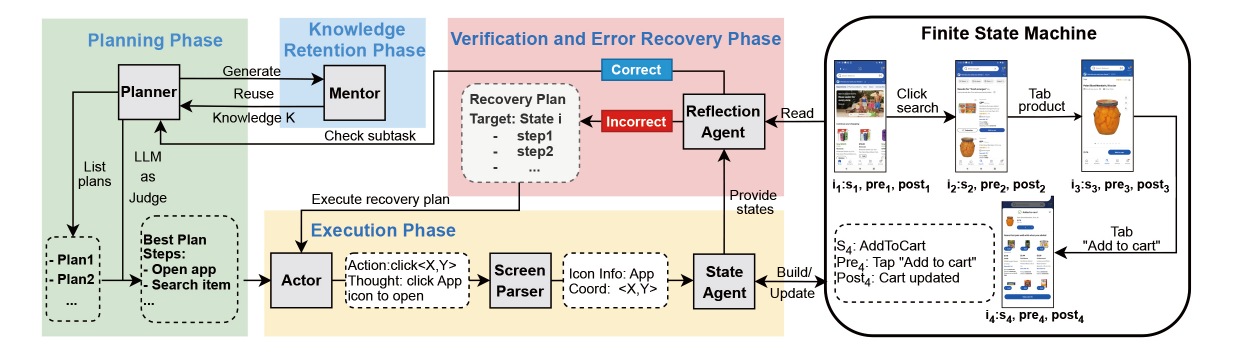}
  \caption{An overview of \tool. The Planner, Actor, Screen Parser, StateAgent, and Reflection Agent are involved in the main agent loop for each task, while Mentor contributes to updating long-term reusable knowledge across tasks. Decision-making at each step is disentangled into high-level planning by the Planner and low-level actions by the Actor. The State Agent builds FSMs dynamically, and the Reflection Agent verifies the outcome of each action, tracks progress, and provides error recovery.}
  \label{fig:agentSAMA}
\end{figure*}

Figure~\ref{fig:agentSAMA} provides an overview of \tool. \tool employs various agents during the four phases of task execution: 1) {\textbf{\textit{Planning}}}, which generates a plan to fulfill a given task. 2) {\textbf{\textit{Execution}}}, which models app UI transition and user actions (e.g., tap or swipe) as a Finite State Machine (FSM), providing a structured representation of the task execution. 
3) {\textbf{\textit{Verification and Error Recovery}}}, where the Reflection Agent leverages the constructed FSM to recover from any encountered error. 
4) {\textbf{\textit{Knowledge Retention}}}, where the Mentor agent extracts reusable knowledge (i.e., guidance cues, action sequences, FSMs) from the execution history, FSM, and error history for guiding future tasks.

\subsection{Planning Phase}
The planning phase is handled by a {\bf Planner agent}, which takes in the user's high-level task instruction, $u$, and generates a structured plan, $\pi$, that decomposes the task into sequential subtasks. 
The Planner first expands $u$ to infer user intent. Then, the Planner generates $\pi = \left[ (g_1, r_1), (g_2, r_2), \ldots, (g_k, r_k) \right]$ with additional reusable knowledge $K$, which generated by prior task execution. Each $g_i \in \mathcal{G}$ represents a subtask (e.g., ``open profile settings''), and each $r_i \in \mathcal{R}$ is a rationale that explains how the subtask contributes to achieving the overall task $u$. As shown in recent work~\cite{gu2025surveyllmasajudge,mahmud2025enhancingllmcodegeneration,seo2025spioensembleselectivestrategies}, relying on a single-path plan may lead to suboptimal results due to lack of consideration on other diverse strategies. Hence, the Planner agent employs a two-step planning pipeline. In the first step, the Planner generates multiple candidate plans, $\Pi = \{ \pi_1, \pi_2, \ldots, \pi_n \}$ where $n=5$. 
In the second step, we leverage LLM-as-judges and ask the LLM agent to evaluate the candidate plans against an evaluation rubric provided by the benchmark that considers factors such as goal relevance, execution efficiency, robustness, and clarity~\cite{gu2025surveyllmasajudge}. The agent then selects the best candidate plan to guide the task execution.

\subsection{Execution Phase}
The execution phase consists of i) \textbf{Screen Parser}, ii) \textbf{State Agent}, and iii) \textbf{Actor Agent}. The \textbf{Screen Parser} serves as the ``\textit{eyes}'' of \tool. 
For every screen that reflects the current state of the device, the Screen Parser performs OCR-based text detection, icon localization, and screen segmentation to identify clickable elements and invokes the MLLM to generate descriptive captions for visual elements, helping agents interpret ambiguous or icon-based UI components. The Screen Parser extracts screen perception information on current screenshot $s_{i}$ before executing any actions and gets $p_{i} = \left[ (e_1, c_1), (e_2, c_2), \ldots, (e_k, c_k) \right]$, where each $(e_j, c_j)$ represents a pair of a GUI element (e.g., screen texts or icon descriptions) and its coordinate. 
The perception result $p_{i}$ and the current screenshot $s_{i}$ are provided to the \textbf{State Agent} as input.

The \textbf{State Agent constructs and incrementally expands app interactions in real-time as a finite state machine (FSM)~\cite{wagner2006fsm}}—an abstract computational model that represents the system as a set of discrete states and transitions triggered by UI actions. 
This structured representation of the app's navigation flow enables the agent to maintain execution context, detect abnormal transitions, and reason about both past and future states. As illustrated in Figure~\ref{fig:fsm}, the FSM represents the UI transitions during task execution, where each state corresponds to a distinct app screen (i.e., \textbf{a state is represented as the natural language description of the screen generated by the State Agent}), and edges represent transitions triggered by user actions such as taps and swipes, maintaining a structured representation of the app's navigation flow.  

The FSM is defined as 
$\mathcal{M} = (S, A, T, s_0, G)$, 
where $S$ is the set of discovered UI states, $A$ is the action space, $T$ is the transition function, $s_0$ is the initial state, and $G$ denotes the goal state for current subgoals in each iteration during the task execution. 
For every perception result $p_i$, screenshot $i_i$ and subtask $g_i$, \textbf{each generated state node $s_{i}$ contains}:
\begin{enumerate}
    \item a natural language \textit{state description} \( d_{\text{i}} \), summarizing the current screen;
    \item a \textit{description of next-state prediction} \( {d}_{\text{i+1}} \), estimating the next expected screen based on the subtask; and
    \item a pre- and post-condition: $\text{pre}^{i+1}$ and $\text{post}^{i}$.
\end{enumerate}
To reduce \textbf{state explosion and avoid redundant states}, \tool assigns each state a state beacon—a concise semantic label derived from its state description \( d_{\text{i}} \). Before creating a beacon for the new FSM node, \textbf{State Agent} will check the state description and decide whether this node can be matched against previously seen beacons. Existing nodes are reused if a match is found, which reduces state explosion caused by redundant states.
Each transition is defined as \( T(s_i, a_i) = (s_{i+1}, \text{pre}^{i+1}, \text{post}^{i}) \)
, where \( s_i \) is the UI state at step $i$, \( a_i \) is the corresponding action, \( s_{i+1} \) is the resulting state, 
$\text{pre}^{i+1}$ is the precondition of $s_{i+1}$, indicating what must hold before the next state begins, and 
$\text{post}^{i}$
is the postcondition of $s_i$, representing what should hold after achieving the state in step $i$.
The \textbf{Actor Agent} selects the corresponding tool (e.g., the mobile API to perform tapping on a specific coordinate) to perform the action $a_i$ based on the current subtask $g_i$ and the perception information $p_i$ to advance to state $s_{i+1}$. Note that for cross-app tasks, we create and maintain separate FSMs for each app.

\subsection{Verification and Error Recovery Phase}
One major benefit of using an FSM to provide a structured representation of app navigation and task execution is its effectiveness in detecting and recovering from execution errors. By explicitly modeling discrete UI states, transitions triggered by user actions, and expected pre- and post-conditions, the FSM allows the system to detect deviations from the intended task flow. 
Hence, in this phase, the \textbf{Reflection Agent} determines whether an action, $a_i$, was successful by examining the result from the FSM (i.e., the prior state $s_{i}$, the predicted next-state description $d_i$, and $\text{pre}_{i-1}$ and $\text{post}_{i}$), the perception result on the new screen $p_{i+1}$, the screenshots ($i$ and $i+1$), the current subtask $g_{j}$, and the knowledge base $K$ .
The agent then outputs one of the three outcomes: {Success}, { NoChange}, or {Fail} (unexpected transition or violation of the predicted state).

In the cases of NoChange or Fail, the Reflection agent also generates the reason for the failures. Then, the agent activates the recovery mode and uses the FSM to identify a previously verified and stable state, $s_j$ to resume the current subtask $g_i$. Then, the reflection agent constructs a recovery plan, \(\pi_r = [a_{i-1},  a_{i-2} \ldots]\), to revert to a recovery point. When executing the recovery plan, the Reflection Agent verifies each step using the same process as in the normal execution flow. If \tool accumulates repeated failures while executing the recovery plan ($n$=2), the recovery process is terminated and \tool re-invokes the \textbf{Planner Agent} to reassess the current state and generate a revised plan. This fallback mechanism prevents the agent from getting stuck in repetitive recovery loops and ensures that high-level reasoning can reorient execution when low-level strategies fail~\cite{wang2025mobile,chang2025sagallmcontextmanagementvalidation}.

\subsection{Knowledge Retention Phase}
\tool generates various types of process data during task execution, including: 1) the action history, 2) error details, and 3) transition history. As shown in prior studies~\cite{wang2025mobile, Liu_2025}, data from prior task execution can be useful to guide future tasks. Hence, at the end of every task, \textbf{Mentor Agent} analyzes this process data to extract reusable knowledge $K$, which includes: (i) \textit{action sequences}, which are sequential actions annotated with preconditions; and (ii) \textit{guidance cues}, which are natural language tips distilled from prior executions to help with future task reasoning. In addition, Mentor also stores the {FSM} in $K$ to better support further tasks if they share similar functionality or design (e.g., shopping events can be similar between Amazon and Walmart). When a new task begins, \tool checks long-term memory for $K$, and selectively loads it as external context (i.e., part of the prompt) to improve planning, execution efficiency, and robustness.

\section{Evaluation}
\label{sec:eval}

\subsection{Benchmarks and Evaluation Settings}
We evaluate \tool on two sets of benchmarks: (1) fully cross-app tasks: Mobile-Eval-E~\cite{wang2025mobile} and SPA-Bench~\cite{chen2025spabench}, and (2) mostly single-app tasks: AndroidWorld~\cite{rawles2025androidworld}. 

\noindent\textbf{Running cross-app benchmarks on a physical device with human evaluation.} Cross-app tasks, such as \textit{e.g., using Chrome to search for the date of the next Winter Olympics opening ceremony, and then setting a reminder for that date in Calendar}, are more realistic and inherently more complex to plan, execute, and recover from~\cite{Liu_2025}. Thus, Mobile-Eval-E and SPA-Bench provide a more rigorous test of GUI agents' capabilities (Table~\ref{tab:benchmark-detail}). To run these two benchmarks, we deploy \tool on a physical Google Pixel 7 Pro, controlled via Android Debug Bridge (ADB). Following the setup in prior work~\cite{wang2025mobile}, we record execution traces and perform human evaluation based on the rubrics defined by Mobile-Eval-E. 
Mobile-Eval-E is composed of 25 manually designed tasks across 15 apps. It has the highest complexity (i.e., requires significantly more actions per task) compared to other existing benchmarks~\cite{wang2025mobile}, and requires multi-app interactions in 76\% of the tasks. 
SPA-Bench provides 40 cross-app tasks, 20 in English and 20 in Chinese, covering both system and third-party apps. For our evaluation, we focused on the 20 English cross-app tasks.

\noindent\textbf{Running single-app benchmark in an emulator with automated evaluation.} To complement the real-device experiments, and enable large-scale automated evaluation, we also evaluate \tool using AndroidWorld. It provides 116 task templates across 20 apps in an emulator environment and supports automated evaluation. Unlike the benchmarks mentioned above, around 90\% of AndroidWorld’s tasks are limited to single-app interactions. 

\begin{table}
\centering
\setlength{\tabcolsep}{1mm}
\scalebox{0.88}{
\begin{tabular}{l rrr}
\toprule
\textbf{Metric} &\textbf{Mobile-Eval-E} & \textbf{SPA-Bench} \\
\midrule
\#Tasks                & 25   & 20  \\
\#Multi-App Tasks       & 19   & 20  \\
\#Apps                & 15   & 25  \\
Avg \# Actions        & 14.56 & 13.10 \\
Total \# Actions      & 364  & 262 \\
\bottomrule
\end{tabular}}\caption{An overview of the cross-app benchmarks, showing the number of tasks, multi-apps, apps, and actions.}

\label{tab:benchmark-detail}
\end{table}

\subsection{Evaluation Metrics} 
\noindent {\bf Cross-app benchmarks.}
App tasks are inherently open-ended and do not always allow a strict binary success criterion~\cite{wang2025mobile}. 
Therefore, following prior work~\cite{wang2025mobile, Liu_2025, wang2024mobile_agent_v2, wang2025_mobile_agent_v}, we rely on human evaluation using the task-specific rubrics provided by Mobile-Eval-E. We then manually adapt the rubrics for SPA-Bench. The first two authors assess the task outcomes separately. In the event of disagreement, the third author serves as the deciding vote to resolve any conflicts. 
We evaluate \tool using five metrics. 

\begin{itemize}[ leftmargin=12pt]
    \item \textbf{Success Rate (SR)} measures the percentage of tasks that are completed successfully~\cite{chen2025spabench}.
    \item \textbf{Satisfaction Score (SS)} quantifies the proportion of rubrics fulfilled for each task, providing a finer-grained view of partial task completion.
    \item \textbf{Action Accuracy (AA)} captures how closely the agent’s executed actions align with the human reference trajectory, computed as the ratio of correctly matched steps.
    \item \textbf{Termination Rate (TR)} reflects the percentage of tasks that terminate unsuccessfully. Following the definition in Mobile-Agent-E~\cite{wang2025mobile}, this includes exits to the home screen, closing the app, or entering unrecoverable states before task completion.
    \item \textbf{Recovery Success (RS)} evaluates the performance of GUI agent's recovery ability, and is defined as the percentage of failed subtasks that are successfully recovered.
\end{itemize}

\noindent {\bf Single-app benchmark.}
AndroidWorld includes an automated evaluation script that verifies task correctness. We also use \textbf{Success Rate (RS)} as the primary metric, where a task is considered successful only if the final UI state precisely matches the expected outcome.

\begin{table}
\centering
\scalebox{0.7}{
\begin{tabular}{l c c c c c}
\toprule
\textbf{MLLM Agent} & SS\% & AA\% & TR\% & SR\% & RS\% \\
\midrule
\multicolumn{6}{c}{\textit{\textbf{Mobile-Eval-E}}}\\
AppAgent$^{*}$        & 25.20 & 60.70 & 96.00 & --    & --    \\
Mobile-Agent$^{*}$  & 45.50 & 69.80 & 68.00 & --    & --    \\
Mobile-Agent-v2$^{*}$       & 53.00 & 73.30 & 52.00 & --    & --    \\
Mobile-Agent-E+Evo  & 78.97 & 76.65 & 24.00 & 72.00 & 67.34 \\
\textbf{Agent-SAMA}            & \textbf{86.15} & \textbf{83.24} & \textbf{16.00} & \textbf{84.00} & \textbf{71.88} \\

\midrule 
\multicolumn{6}{c}{\textit{\textbf{SPA-Bench}}}\\

Mobile-Agent-E+Evo & 80.30 & 77.86 & 25.00 & 75.00 & 52.86 \\
\textbf{Agent-SAMA}           & \textbf{88.64} & \textbf{84.35} & \textbf{20.00} & \textbf{80.00} & \textbf{66.67} \\

\bottomrule
\end{tabular}}
\caption{Comparison of evaluation metrics on Mobile-Eval-E and SPA-Bench. Higher is better except for Termination Rate (TR). We adopted the results from prior work~\cite{wang2025mobile} for the agents marked with $^{*}$.}

\label{tab:comparison_mobile_eval}
\end{table}

\subsection{Experiment Settings}

\begin{table}
\centering
\setlength{\tabcolsep}{25pt}

\scalebox{0.7}{

\begin{tabular}{l r}
\toprule
\textbf{MLLM Agent} & \textbf{Success Rate (\%)} \\
\midrule
GPT-4 Turbo$^{*}$ & 30.6 \\
GPT-4o$^{*}$ & 34.5 \\
GPT-4o+UGround$^{*}$ & 44.0 \\
GPT-4o+Aria-UI$^{*}$ & 44.8 \\
Mobile-Agent-E & 45.7 \\
UI-TARS$^{*}$ & 46.6 \\
Mobile-Agent-E+Evo & 53.4 \\
AgentS2$^{*}$ & 54.3 \\
V-Droid$^{*}$ & 59.5 \\
\textbf{\tool} & 63.7 \\
\bottomrule
\end{tabular}}\caption{Task Success Rate (\%) on AndroidWorld benchmark. We adopted the results from V-Droid~\cite{dai2025V_Droid} for the agents marked with $^{*}$.}
\label{tab:ad}
\end{table}

\phead{\textbf{Backbone MLLMs.}}
We use GPT-4o-2024-11-20~\cite{openai2024gpt4o} as the backbone MLLM in our experiments with a temperature of 0 to reduce variations. GPT-4o is a state-of-the-art MLLM capable of processing both text and images, making it suitable for complex mobile interaction tasks. GPT-4o offers strong performance with particularly low latency and efficient image-text alignment.
\phead{Screen Parser Implementation.}
Our Screen Parser is implemented in alignment with Mobile-Agent-E~\cite{wang2025mobile}. For OCR detection and recognition, we adopt DBNet~\cite{liao2020_DBNet} and ConvNextViT-document from ModelScope, respectively. Icon grounding is performed using GroundingDINO~\cite{liu2024groundingdinomarryingdino}, while Qwen-VL-Plus~\cite{bai2023_qwen} is used to generate textual captions for each detected icon crop.

\subsection{Experiment and Results}
 
\textbf{Performance of \tool.} Table~\ref{tab:comparison_mobile_eval} compares \tool with prior baselines, including AppAgent~\cite{yang2023appagent} and Mobile-Agent series~\cite{wang2024mobile_agent,wang2024mobile_agent_v2,wang2025mobile} on the Mobile-Eval-E benchmarks across five evaluation metrics. We re-ran Mobile-Agent-E+Evo as our main baseline because it is the only prior agent with long-term memory and knowledge reuse, which are key features for fair comparison with \tool. Other baselines lack memory or structured agent frameworks and have been consistently outperformed. We then compare \tool with Mobile-Agent-E+Evo on the SPA-Bench. 
To ensure consistency, we re-ran Mobile-Agent-E+Evo five times using the same MLLM as \tool, with two authors evaluating the results independently. Finally, we report the average.

\begin{table}
\centering
\scalebox{0.71}{
\setlength{\tabcolsep}{1mm}
\begin{tabular}{llccc}
\toprule
\textbf{Benchmark} & \textbf{Metric} & \textbf{GPT-4o} & \textbf{Claude 3.5} & \textbf{Gemini 1.5 Pro} \\
\midrule
\multirow{4}{*}{M-Eval-E} 
& SS & 168 / 195 (86.15\%) & 158 / 195 (81.03\%) & 137 / 195 (70.25\%) \\
& AA    & 303 / 364 (83.24\%) & 289 / 364 (79.40\%) & 241/ 364 (66.20\%) \\
& TR   & 4 / 25 (16.00\%)     & 5 / 25 (20.00\%)   & 8 / 25 (32.00\%) \\
& SR       & 21 / 25 (84.00\%)    & 20 / 25 (75.00\%)   &  17/ 25 (68.00\%) \\
\midrule
\multirow{4}{*}{SPA-Bench} 
& SS & 117 / 132 (88.64\%) & 108 / 132 (81.82\%) & 98 / 132 (74.24\%) \\
& AA    & 221 / 262 (84.35\%) & 215 / 262 (82.06\%) & 181 / 262 (69.08\%) \\
& TR   & 4 / 20 (20.00\%)   & 4 / 20 (20.00\%)   & 7 / 20 (35.00\%) \\
& SR      & 16 / 20 (80.00\%)  & 16 / 20 (80.00\%)   & 12 / 20 (60.00\%) \\
\bottomrule
\end{tabular}}\caption{Comparison of \tool's performance across Mobile-Eval-E and SPA-Bench using different backbones. }
\label{tab:backbone}
\end{table}

Across both benchmarks, \tool improves all evaluation metrics compared to the baseline by 4.53\% to over 13.00\%, indicating \tool is able to finish more cross-app tasks with higher accuracy and quality. 
On Mobile-Eval-E, \tool achieves a 7.18\% improvement in Satisfaction Score (SS), 6.59\% in Action Accuracy (AA), 8\% in Termination Rate (TR), 4.53\% in Recovery Success(RS), and a notable 12\% in Success Rate (SR) in comparison with the strongest baseline Mobile-Agent-E+Evo. 
We see similar improvements on SPA-Bench, where \tool achieves an 8.33\% improvement in SS, 6.49\% in AA, 5\% in TR, a significant 13.81\% in RS, and a 5\% in SR. 
These results indicate that \tool's FSM-based architecture contributes significantly to robust, accurate, and resilient task execution in GUI agents.

\begin{table*}[h]
\centering
\scalebox{0.7}{\setlength{\tabcolsep}{1.7mm}
\begin{tabular}{cccc|ccc|ccc}
\toprule
\multicolumn{4}{c|}{\textbf{Ablation Setting}} & \multicolumn{3}{c|}{\textbf{Mobile-Eval-E}} & \multicolumn{3}{c}{\textbf{SPA-Bench}} \\
\textbf{Planning} & \textbf{Multi-plans} & \textbf{Pre/Post conditions} & \textbf{Mentor} & \textbf{SS (\%)} & \textbf{AA (\%)} & \textbf{SR (\%)} & \textbf{SS (\%)} & \textbf{AA (\%)} & \textbf{SR (\%)} \\
\midrule
 & & \checkmark & \checkmark & 61.54 & 62.91 & 52.00 & 56.06 & 54.96 & 45.00 \\
 \checkmark &  & \checkmark & \checkmark & 72.31 & 67.22& 68.00 & 71.97 & 69.47 & 60.00 \\
\checkmark&\checkmark &  & \checkmark & 82.68 & 78.25 & 72.00 & 81.77 & 79.44 & 70.00 \\
\checkmark&\checkmark & \checkmark &  & 73.85 & 73.08 & 68.00 & 78.79 & 74.05 & 70.00 \\
\checkmark&\checkmark & \checkmark & \checkmark & \textbf{86.15} & \textbf{83.24} & \textbf{84.00} & \textbf{88.64} & \textbf{84.35} & \textbf{80.00} \\
\bottomrule
\end{tabular}
}\caption{The results of the ablation study when removing 1) Planner agent (i.e., no planning at all), 2) Selecting the best plan across multiple plans, 3) Pre and Post conditions in State Agent, and 4) Mentor agent. The last row shows results from the full version of \tool. }

\label{tab:ablation}
\end{table*}
Table~\ref{tab:ad} presents the result of \tool on the AndroidWorld benchmark. The table contains popular baselines such as AgentS2~\cite{agashe2025agents2}, V-Droid~\cite{dai2025V_Droid}, and Mobile-Agent-E~\cite{wang2025mobile} and its Evo version. \tool demonstrates competitive performance, achieving a success rate of 63.7\%, outperforming most existing mobile agent baselines. While AndroidWorld tasks are generally shorter than those in Mobile-Eval-E and SPA-Bench, the benchmark enforces strict step limits and low fault tolerance~\cite{rawles2025androidworld}, making recovery more costly. Despite this, \tool maintains strong performance through stable planning and execution.

\phead{\textbf{Sensitivity of the results when changing the backbone MLLM.}} Table~\ref{tab:backbone} shows \tool's overall performance across three widely used MLLMs: GPT-4o~\cite{openai2024gpt4o}, Claude-3.5-Sonnet~\cite{anthropic2024claude}, and Gemini 1.5 Pro~\cite{gemini2024} on both Mobile-Eval-E and SPA-Bench benchmarks. 
The backbone MLLM has a large impact on the performance of \tool, with GPT4o achieves the highest scores, followed by Claude 3.5, with Gemini 1.5 Pro performing the lowest. 
This trend is consistent across all three major evaluation metrics, suggesting that stronger MLLMs improve GUI agent's performance. Nevertheless, when using a weaker MLLM like Claude 3.5, \tool still outperforms Mobile-Agent-E Evo (GPT-4o version), demonstrating the effectiveness of \tool. 

A notable comparison is on error occurrence and recovery, as \tool uses FSM to assist in reasoning and error recovery. We find that {\tool encounters notably fewer less errors} during execution compared to Mobile-Agent-E + Evo (32 vs. 49 on Mobile-Eval-E and 63 vs. 70 on SPA-Bench) {with a higher recovery rate} (4.53\% and 13.81\% higher). The findings show that \tool's structured FSM representation improves the agent’s ability to make correct decisions, while also enhancing its capacity to recover from errors. 

\textbf{Ablation Studies.} Table~\ref{tab:ablation} presents the results of our ablation study on the contributions of four key components in \tool: 1) the entire planning (removing the Planner Agent completely), 2) LLM-as-judges~\cite{gu2025surveyllmasajudge} to select the best plan among five generated plans, 3) pre- and post-conditions in the State Agent, and 4) the entire knowledge retention (removing the Mentor agent completely). 
We observe that each component contributes significantly to \tool's performance, \textit{\textbf{and their combination leads to more substantial improvements}}. 
Removing each component reduces the performance metrics significantly, with planning having the largest impact (e.g., SR decreased from 84\% to 52\% in Mobile-Eval-E and from 80\% to 45\% on SPA-Bench), followed by multi-plan selection, Mentor agent, and then pre- and post-conditions. 
Importantly, when all components are combined, \tool achieves the highest performance across all metrics. This demonstrates that these components complement and reinforce one another: effective planning helps the system select better paths, selecting the best plan from multiple ones may help reduce the effects of having excessive states on subtask planning or re-planning during error recovery, pre- and post-conditions enable better validation and recovery, and the Mentor allows learning across tasks. Overall, the full integration leads to an effective state-aware GUI agent framework.

\section{Limitations}
\label{sec:limitation}
\textbf{Benchmark Coverage. }
Our evaluation is limited to Mobile-Eval-E and SPA-Bench. Although these are the state-of-the-art benchmarks with the highest difficulties, cross-app tasks, and represent realistic usage scenarios, they cannot fully reflect all interactions. Some interactions (e.g., dynamic content, external interruptions like third-party ads) may even trigger unpredictable states. Future studies are needed to evaluate \tool on more diverse scenarios and tasks. 

\noindent\textbf{Baseline Comparison.}
We included the Mobile Agent series as our primary baseline, given its strong performance on long-horizon mobile tasks. To provide border comparison, we also included additional agent frameworks such as AppAgent~\cite{yang2023appagent} and AgentS2~\cite{agashe2025agents2}. Notably, the results we obtained from re-running Mobile-Agent-E+Evo differ from those reported in its original paper. To migrate variation across runs, we re-ran five times with the same settings, and two authors evaluated the results independently. Finally, we reported the average. We also observe similar results across runs. 

\section{Conclusion}
\label{sec:conclusion}

In this paper, we introduced \tool, a state-aware mobile GUI agent that models app navigation and task execution as a Finite State Machine (FSM). By design, specialized agents for various phases of task execution—planning, execution, verification, and recovery, and knowledge retention—\tool provides a structured representation and effective framework for mobile task automation. 
Our evaluation on two real-world cross-app benchmarks(Mobile-Eval-E and SPA-Bench) and one mostly single-app benchmark(AndroidWorld) shows that \tool achieves up to 12\% improvement in Success Rate, and 13.8\% in Recovery Success compared to the baseline Mobile-Agent-E+Evo~\cite{wang2025mobile} on cross-app benchmarks, and it also achieves a high success rate of 63.7\% on AndroidWorld. 
These results highlight the value of FSM-based modeling for improving agent planning and error recovery in complex mobile environments. 

{
\bibliography{aaai2026}
}
\onecolumn
\newpage
\appendix
\section{Technical Appendices}
\subsection{Trajectory Comparison Example with Previous SOTA }
\begin{figure}[p] 
  \centering
  \includegraphics[width=0.64\paperwidth,height=1\paperheight,keepaspectratio]{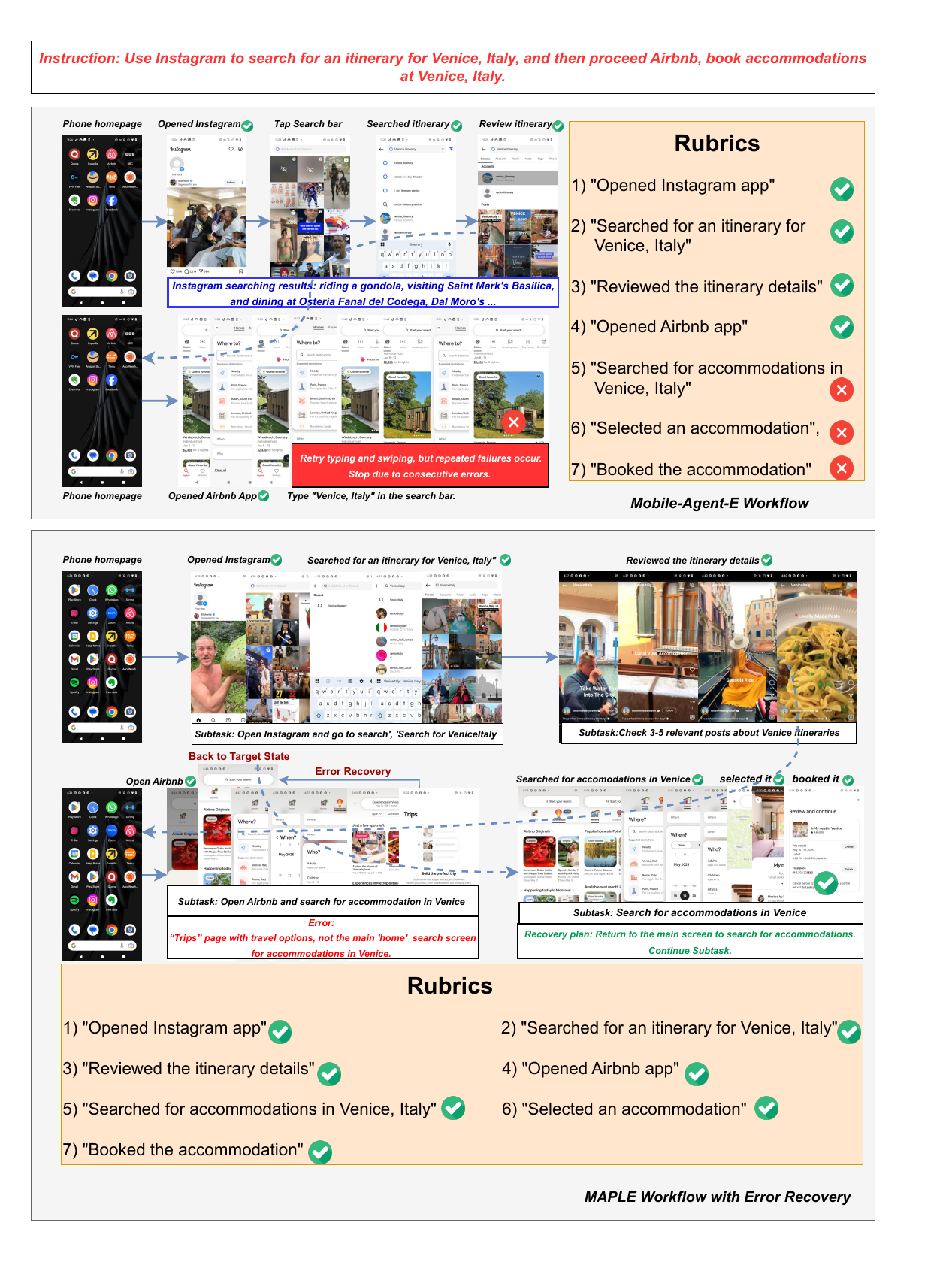}
  \caption{Comparison between the previous state-of-the-art, Mobile-Agent-E~\cite{wang2025mobile}, and \tool.}
  \label{fig:appendix_workflow}
\end{figure}

Figure~\ref{fig:appendix_workflow}  presents the full trajectory of the task in SPA-Bench~\cite{chen2025spabench}, comparing the previous state-of-the-art, Mobile-Agent-E~\cite{wang2025mobile}, and our proposed \tool. \tool fulfills all rubrics and stops where the app completes the task.
Figure~\ref{fig:appendix_workflow}  presents the full trajectory of the task in SPA-Bench~\cite{chen2025spabench}, comparing the previous state-of-the-art, Mobile-Agent-E~\cite{wang2025mobile}, and our proposed \tool. \tool fulfills all rubrics and stops where the app completes the task.

\subsection{Error Recovery with Recovery Plan}
\label{app:B}

This subsection presents a real-world example demonstrating how the agent utilizes a recovery plan to resume execution after encountering an error during task execution. Specifically, the plan is designed to return to a previously stable state and reattempt the failed subtask with improved reliability.

\vspace{0.5em}

\begin{tcolorbox}[title=Recovery Goal, colback=gray!5!white, colframe=gray!75!black]
Return to the search results page in the Maps app to reattempt the address selection and copy process.
\end{tcolorbox}

\begin{tcolorbox}[title=Recovery Target State ID, colback=gray!5!white, colframe=gray!75!black]
e66d543fcc30c83110b07215f295f986
\end{tcolorbox}

\begin{tcolorbox}[title=Recovery Plan (JSON), colback=gray!5!white, colframe=gray!75!black]
\lstset{basicstyle=\ttfamily\small}
\begin{lstlisting}
{
  "thought": "The selected state represents the search results page in the Maps app, where the address of the Alo Yoga store is visible. This state is ideal for recovery because it allows the agent to reattempt the process of selecting and copying the address. The error likely occurred due to an incorrect tap action, so returning to this state provides a clean slate to retry the task. The plan involves navigating back to this state by ensuring the Maps app is active and the search results page is displayed.",
  "plan": [
    "Ensure the Maps app is active and in the foreground.",
    "Navigate back to the search results page if not already visible."
  ],
  "current_subtask": "Ensure the Maps app is active and in the foreground."
}
\end{lstlisting}
\end{tcolorbox}

\subsection{All used Tasks in SPA-Bench}
Table~\ref{tab:task-queries} presents the input queries, involved Task ID, App names, and scenarios for all SPA-Bench ~\cite{chen2025spabench} tasks we used in evaluation. The complete list of rubrics and human reference operation sequences is provided in the supplementary material.

{
\renewcommand{\tabularxcolumn}[1]{>{\centering\arraybackslash}m{#1}}
\begin{table}[htbp]
\tiny
\centering
\caption{All task queries in SPA-Bench}
\label{tab:task-queries}
\begin{tabularx}{\textwidth}{p{2.5cm} p{2.8cm} p{2.5cm} X}
\toprule
\textbf{Scenario} & \textbf{Task ID} & \textbf{APPs} & \textbf{Input Query} \\
\midrule

\multirow{5}{=}{Device utilities} 
& General\_Tool\_0\_ENG & Google Play Store, Setting & Open Google Play Store, uninstall the Alibaba.com app, then go to Settings and verify if the app is still listed under app resources. \\
\cline{2-4} 
& General\_Tool\_2\_ENG & Clock, Setting & In the Settings app, enable 'Data Saver' mode. Open the Clock app and set an alarm for 6:00 AM. \\
\cline{2-4} 
& Media\_Entertainment\_2\_ENG & Clock, YouTube & Search for a relaxing soundscape video on YouTube, use the Clock app to set a timer for 3 hours, then go back to YouTube and play the video. \\
\cline{2-4} 
& Multi\_Apps\_2\_ENG & Triller, Google Play Store, Setting & First, install the Triller app from the Google Play Store. After the installation, open the Triller app, navigate to the Setting app to check current battery status, reopen the Triller app. \\
\cline{2-4} 
& Multi\_Apps\_3\_ENG & Clock, WhatsApp, Zoom & Arrange a business meeting using Zoom, copy the sharing text, go to WhatsApp, send the copied text to a contact, set an alarm using the Clock app at the meeting time. \\
\midrule
\multirow{2}{=}{Information Research}
& General\_Tool\_1\_ENG & Keep Notes, LinkedIn & Use the LinkedIn app to search for a customer service representative position. Select a job, open Keep Notes, create a new note, record the company's name, and set the note's title to 'customer service representative'. \\
\cline{2-4} 
& Information\_Management\_1\_ENG &   Chrome, Calendar & Using Chrome, search for the date of the next Winter Olympics opening ceremony and then set a reminder for that date in your Calendar. \\
\cline{2-4}  
& Media\_Entertainment\_1\_ENG & Google Play Store, Chrome & Utilize Chrome to research different Recipe Organizer apps, and then proceed to Google Play Store, download one of your choice. \\
\cline{2-4} 
& Social\_Sharing\_1\_EN & BBC News, Gmail & Use the BBC News app to search for Artificial Intelligence news, read an article, share it via Gmail, send to xxxxxx.2025@gmail.com. \\
\cline{2-4} 
& Media\_Entertainment\_0\_ENG & YouTube, Google Play Store & Watch a YouTube video about fitness tracking app recommendations, check the video's description for the suggested apps, then use Google Play Store to download one of the suggested apps.\\
\midrule
\multirow{2}{=}{whats trending}
& Information\_Management\_0\_ENG & Facebook, Setting & Open Facebook, search for tropical pictures, save one picture, go to the Wallpaper section in the Settings app, and set the saved picture as your wallpaper \\
\cline{2-4} 
& Information\_Management\_2\_ENG & Spotify, Chrome & Open Chrome, search for the top Country songs of 2023, identify a song from the search results, then switch to Spotify and add to your playlist.\\
\cline{2-4} 
& Multi\_Apps\_1\_ENG & Clock, Chrome, Instagram & Choosing a horror film using Chrome, sending an invitation to one of your friends via Instagram, and setting a reminder in the Clock app for 8:35 PM on Sunday.\\
\cline{2-4} 
& Social\_Sharing\_0\_ENG & X, Facebook & Use the social media platform X to post a photo, copy the link to your post, then open Facebook and send the link to a friend.\\
\cline{2-4} 
& Social\_Sharing\_2\_ENG & Spotify, Facebook & Listen to a Reggaeton album on Spotify, then share the album's name with a friend on Facebook\\
\midrule

\multirow{2}{=}{online shopping \& travel}
& Multi\_Apps\_0\_ENG & Quora, eBay, Chrome & Utilize Chrome to search for a biography book, then use Quora to read reviews about the book, and finally add the book to watchlist on eBay \\
\cline{2-4} 
& Multi\_Apps\_4\_ENG & AccuWeather, Evernote, Expedia & Open Chrome, search for the top Country songs of 2023, identify a song from the search results, then switch to Spotify and add that song to your playlist.\\
\cline{2-4} 
& Web\_Shopping\_0\_ENG & eBay, Facebook & Search for 'Circe by Madeline Miller' on Facebook, read one of the posts, head over to eBay, search the book, add it to watchlist.\\
\cline{2-4} 
& Web\_Shopping\_1\_ENG & Amazon, Temu & Investigate the prices for Catan board game across Amazon and Temu, then proceed to add the cheaper option into your cart.\\
\cline{2-4} 
& Web\_Shopping\_2\_ENG & Instagram, Airbnb & Use Instagram to search for an itinerary for Venice, Italy, and then proceed Airbnb, book accommodations at Venice, Italy.\\

\bottomrule
\end{tabularx}
\end{table}
}

\subsection{Task Evaluation Rubrics and Operations}
\label{app:rubrics}
This subsection shows a cross-app task from SPA-Bench ~\cite{chen2025spabench}, involving Instagram and Airbnb. The task details, evaluation rubrics, and reference operations are presented in JSON format. The rubrics outline key steps like searching for an itinerary and booking accommodations, while the operations specify concrete user actions for evaluation.

\begin{tcolorbox}[title=Task details (JSON), colback=gray!5!white, colframe=gray!75!black]
\lstset{basicstyle=\ttfamily\small}
\begin{lstlisting}
{
  "task_id": "Web_Shopping_2_ENG",
  "instruction": "Use Instagram to search for an itinerary for Venice, Italy, and then proceed Airbnb, book accommodations at Venice, Italy."
  "type": "multi_app",
  "apps": [
            "Airbnb",
            "Instagram"
          ]
}
\end{lstlisting}
\end{tcolorbox}
\begin{tcolorbox}[title=Task Evaluation Rubrics (JSON), colback=gray!5!white, colframe=gray!75!black]
\lstset{basicstyle=\ttfamily\small}
\begin{lstlisting}
{
  "rubrics": [
    "Opened Instagram app",
    "Searched for an itinerary for Venice, Italy",
    "Reviewed the itinerary details",
    "Opened Airbnb app",
    "Searched for accommodations in Venice, Italy",
    "Selected an accommodation",
    "Booked the accommodation"
  ]
}
\end{lstlisting}
\end{tcolorbox}

\begin{tcolorbox}[title=Task Evaluation Operations (JSON), colback=gray!5!white, colframe=gray!75!black]
\lstset{basicstyle=\ttfamily\small}
\begin{lstlisting}
{
  "human_reference_operations": [
    "open Instagram app",
    "tap on the search bar",
    "type 'Venice Italy itinerary'",
    "tap enter",
    "scroll through results and select an itinerary post",
    "review the itinerary details",
    "press home button",
    "open Airbnb app",
    "tap on the search bar",
    "type 'Venice, Italy'",
    "tap enter",
    "scroll through accommodation options",
    "select an accommodation",
    "tap 'Reserve' or 'Book'",
    "enter booking details and confirm the reservation"           
  ]
}
\end{lstlisting}
\end{tcolorbox}

\subsection{Actor Agent Tool Box}
\label{app:tools}
Table~\ref{tab:atomic-operations} outlines the fundamental action set used by the \textbf{Actor Agent} to interact with mobile user interfaces. These actions represent low-level actions such as tapping, typing, swiping, and app navigation, which are composed to execute higher-level tasks across applications. The design of this action space is informed by prior work on mobile agents ~\cite{wang2025mobile}.
\renewcommand{\tabularxcolumn}[1]{>{\raggedright\arraybackslash}p{#1}}

\begin{table}[htbp]
\centering
\footnotesize
\caption{Actor Agent Tool Box.}
\label{tab:atomic-operations}
\renewcommand{\arraystretch}{1.3}

\begin{tabularx}{\linewidth}{p{3.5cm} X}
\toprule
\textbf{Action} & \textbf{Description} \\
\midrule
\textit{Tap(x, y)} & Simulate a tap gesture at the screen coordinates \((x, y)\). \\
\textit{Type(text)} & Input the specified \texttt{text} string into a visible text field. \\
\textit{Enter()} & Emulate pressing the enter key, typically after typing a query. \\
\textit{Back()} & Navigate one step back in the current app’s screen history. \\
\textit{Open\_App(app\_name)} & Launch the application identified by \texttt{app\_name} if it is currently visible on the Home or App screen. \\
\textit{Swipe($x_1$, $y_1$, $x_2$, $y_2$)} & Perform a swipe gesture from start point \((x_1, y_1)\) to end point \((x_2, y_2)\). This can be used to scroll through content vertically or horizontally. \\
\textit{Switch\_App()} & Trigger the app switcher interface to access running applications. \\
\textit{Home()} & Return to the device's home screen. \\
\textit{Wait()} & Pause the interaction for 10 seconds to allow background processes or loading to complete. \\
\bottomrule
\end{tabularx}
\end{table}

\subsection{Agent Prompts}

\begin{tcolorbox}[title=State Agent Prompt, colback=gray!5, colframe=gray!75!black, fontupper=\ttfamily\small, breakable]
\#\#\# Current Screen State Analysis \#\#\#
Using the provided screenshot and its perception data, describe the current screen state.

\#\#\# Your Output Format \#\#\#
Respond in the exact format below:

\#\#\# State Description \#\#\#
<Brief natural language summary of the screen>

\#\#\# Predicted Next State \#\#\#
Based on the user instruction, the global plan, and the current subgoal, predict what the screen will display once the next subgoal is successfully achieved.

\#\#\# App Inference \#\#\#
Guess the name of the app currently being used based on the screen content. If the screen shows the Home screen, App Drawer, or any system UI with no app actively open, respond with 'System Operation'.\\
Note that: ONLY RETURN THE APP NAME here, no more other words. Check the previous beacon history to avoid having different variants of the same app.

\#\#\# State Beacon \#\#\#
Use a very strict format such as:
\begin{itemize}[leftmargin=1.5em, label=--]
    \item Homepage of \textless AppName\textgreater
    \item Search Page of \textless AppName\textgreater
    \item Item Detail Page in \textless AppName\textgreater
    \item Notes List in \textless AppName\textgreater
    \item Settings Page of \textless AppName\textgreater
\end{itemize}
Use consistent naming. If a similar beacon has appeared before, reuse it exactly.\\

Previously seen beacon history:

\#\#\# Post-condition of Current State \#\#\#
After completing the current subgoal, what should be true in the app or screen? State it clearly.

\#\#\# Pre-condition of Next State \#\#\#
What needs to be true before the next subgoal can begin execution?
\end{tcolorbox}

\end{document}